\documentclass[10pt,twocolumn,letterpaper]{article}

\usepackage{cvpr}              

\usepackage{microtype}
\usepackage{graphicx}
\usepackage{booktabs} 

\usepackage{amsmath}
\usepackage{amssymb}
\usepackage{mathtools}
\usepackage{amsthm}
\usepackage{makecell}
\usepackage{microtype}      
\usepackage{color}          
\usepackage{bm}
\usepackage{multirow, booktabs}
\definecolor{Graylight}{gray}{0.9}
\usepackage{comment}
\usepackage{xcolor,colortbl}
\newcommand{\dw}[1]{\textcolor{red}{\small{dw: #1}}}
\newcommand{\lw}[1]{\textcolor{blue}{\small{lw: #1}}} 
\newcommand{\ml}[1]{\textcolor{purple}{\small{ml: #1}}} 
\newcommand{\yx}[1]{\textcolor{violet}{\small{yx: #1}}} 
\newcommand{\red}[1]{\textcolor{red}{#1}}
\newcommand{\blue}[1]{\textcolor{black}{#1}}

\theoremstyle{plain}

\theoremstyle{definition}

\theoremstyle{remark}

\usepackage[textsize=tiny]{todonotes}

\newcommand{\R}{\mathbb{R}}

\definecolor{cvprblue}{rgb}{0.21,0.49,0.74}
\usepackage[pagebackref,breaklinks,colorlinks,citecolor=cvprblue]{hyperref}

\usepackage[capitalize,noabbrev]{cleveref}

\def\paperID{244} 
\def\confName{3DV\xspace}
\def\confYear{2024\xspace}

\title{PathFusion: Path-Consistent Lidar-Camera Deep Feature Fusion}

\author{
Lemeng Wu\textsuperscript{1} ~~~~~~~~Dilin Wang\textsuperscript{1} ~~~~~~~~Meng Li\textsuperscript{2} ~~~~~~~~Yunyang Xiong\textsuperscript{1}  
\\Raghuraman Krishnamoorthi\textsuperscript{1} ~~~~Qiang Liu\textsuperscript{3} ~~~~Vikas Chandra\textsuperscript{1} \\
\textsuperscript{1} Meta Reality Labs~~
\textsuperscript{2} Peking University~~~~
\textsuperscript{3} University of Texas at Austin \hspace{10pt}  \\
{\tt \footnotesize \{lmwu, wdilin, yunyang, raghuraman, vchandra\}@meta.com, } 
\tt \footnotesize {meng.li@pku.edu.cn, lqiang@cs.utexas.edu}
}

\begin{document}
\maketitle

\begin{abstract}
Fusing 3D LiDAR features with 2D camera features is a promising technique for enhancing the accuracy of 3D detection, thanks to their complementary physical properties.
While most of the existing methods focus on directly fusing camera features with raw LiDAR point clouds or shallow-level 3D features,
it is observed that directly combining 2D and 3D features in deeper layers actually leads to a decrease in accuracy due to feature misalignment.
The misalignment, which stems from the aggregation of features learned from large receptive fields, becomes increasingly more severe as we delve into deeper layers.
In this paper, we propose PathFusion as a solution to enable the alignment of semantically coherent LiDAR-camera deep feature fusion. 
PathFusion introduces a path consistency loss at multiple stages within the network, 
encouraging the 2D backbone and its fusion path to transform 2D features in a way that aligns semantically with the transformation of the 3D backbone. This ensures semantic consistency between 2D and 3D features, even in deeper layers, and amplifies the usage of the network's learning capacity.
We apply PathFusion to improve a prior-art fusion baseline, Focals Conv, and observe an improvement of over 1.2\% in mAP on the nuScenes test split 
consistently with and without testing-time data augmentations, and moreover, PathFusion also improves KITTI $\text{AP}_{\text{3D}}$ (R11) by about 0.6\% on the moderate level.
\end{abstract}

\section{Introduction}

LiDARs and cameras are widely used in autonomous driving as they offer  complementary information for 3D detection~\citep[e.g.,][]{nuscenes,waymo,kitti}.
While LiDAR point clouds capture better geometry information, they are limited by low resolution due to the power and hardware limitations.
On the other hand, cameras capture dense and colored images, rich in semantic information, but typically lack the shape and depth details required for geometry reasoning.
Consequently, recent methods~\citep{li2022bevformer,pointpainting,liang2022bevfusion,deepfusion} propose fusing 2D and 3D features
from LiDAR point clouds and camera images to enable accurate and robust 3D detection.

The fusion of LiDAR and camera necessitates the projection of features into a unified feature space.
Previous studies have suggested either lifting 2D camera features into 3D space~\citep[e.g.,][]{pointpainting,pointaugmenting,deepfusion,chen2022autoalign,focalsconv-chen}
or aligning both 2D and 3D features within a shared representation space, such as bird's-eye view (BEV)~\citep[e.g.,][]{liu2022bevfusion,liang2022bevfusion}.
\begin{figure*}[t]
    \centering
    \includegraphics[width=\textwidth]{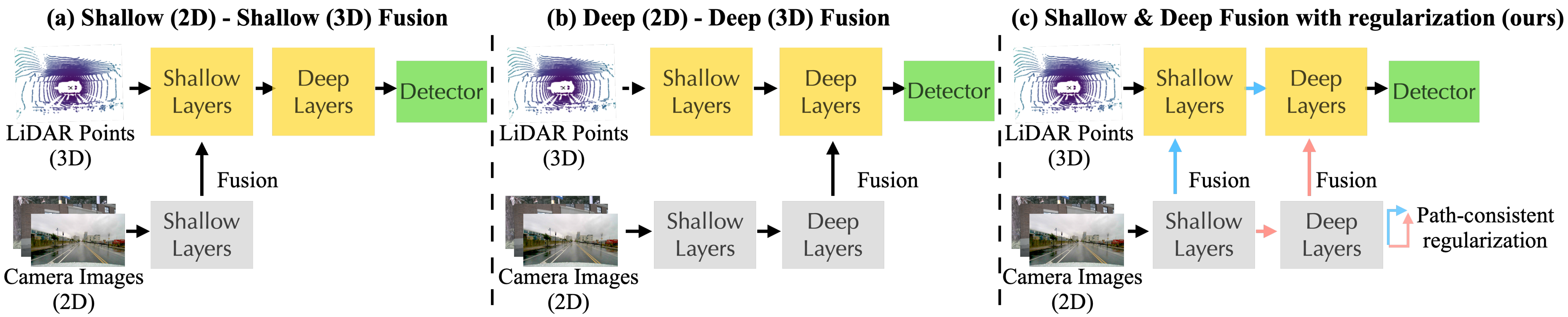}
    \caption{Overview of three different strategies to fuse the camera and LiDAR features:
        (a) Shallow fusion accurately fuses the 2D feature with the shallow 3D feature;
        (b) Deep fusion involves projecting the 2D features into the 3D feature space.;
        (c) Our method introduces the path consistency loss to mitigate the issue of feature misalignment.
    }
    \label{fig:overview}
\end{figure*}
One important question for feature fusion is to select the correct fusion stage, and we illustrate popular choices in Figure~\ref{fig:overview}.
Figure~\ref{fig:overview} (a) showcases shallow fusion~\citep{focalsconv-chen,pointpainting,chen2022autoalignv2}, where camera features are directly fused with raw LiDAR points or shallow LiDAR features.
Although shallow fusion benefits from the alignment facilitated by LiDAR and camera calibration, it forces camera features to pass through multiple modules designed for 3D feature extraction instead of 2D camera processing~\citep{deepfusion}, such as voxelization.
In contrast, deep fusion~\citep{fusionpainting,deepfusion} in Figure~\ref{fig:overview} (b) enables more specialized LiDAR and camera feature processing.
However, due to the typical voxelization of multiple LiDAR points together, followed by aggregation with neighboring voxels prior to fusion,
a single voxel may correspond to camera features derived from a large region (e.g., a union of foreground and background), leading to large ambiguity in feature alignment.
This misalignment tends to have a notable negative impact on the overall network accuracy, often motivating many of the existing works to lean towards opting for shallow fusion~\citep{focalsconv-chen}.

In this paper, we introduce PathFusion, a method designed to enhance the fusion of deep LiDAR and camera features. As depicted in Figure~\ref{fig:overview} (c), our method works by constructing multiple fusion paths connecting 2D and 3D features. Starting from the same 2D camera features, our consistency loss enforces the feature produced by the 2D branch (e.g., the red path in Figure~\ref{fig:overview} (c)) to match the feature produced by the 3D branch (e.g., the blue path in Figure~\ref{fig:overview} (c)). Note that this is different from directly minimizing the difference between 2D and 3D, as they stem from different modalities and cannot be easily matched. Instead, our method traverses the same 2D inputs via the 2D branch and 3D branch separately, and regularizes the 2D branch to learn features that align with the 3D branch. Through simple mathematical induction, it's easy to see the network trained in this manner will yield semantically aligned features across the network.

Meanwhile, in contrast to prior approaches that propose new model architectures~\citep[e.g.,][]{yang2022deepinteraction,bai2022transfusion,deepfusion} or new surrogate tasks~\citep{liu2022bevfusion}, our work offers a fresh perspective by focusing on training regularization, which can be seamlessly integrated into existing methods. In particular, we apply our method to FocalConv~\citep{focalsconv-chen} and expand its shallow fusion into deep fusion, demonstrating consistent improvements in accuracy. Specifically, when compared to FocalConv, PathFusion achieves a 0.62\% improvement in $\text{AP}_{\text{3D}}$ (R11) on the KITTI dataset, along with a 1.2\% increase in mAP and a 0.7\% enhancement in NDS on the nuScenes test set.

\section{Related Work}
\begin{figure*}[!htbp]
    \centering
    \includegraphics[width=0.8\textwidth]{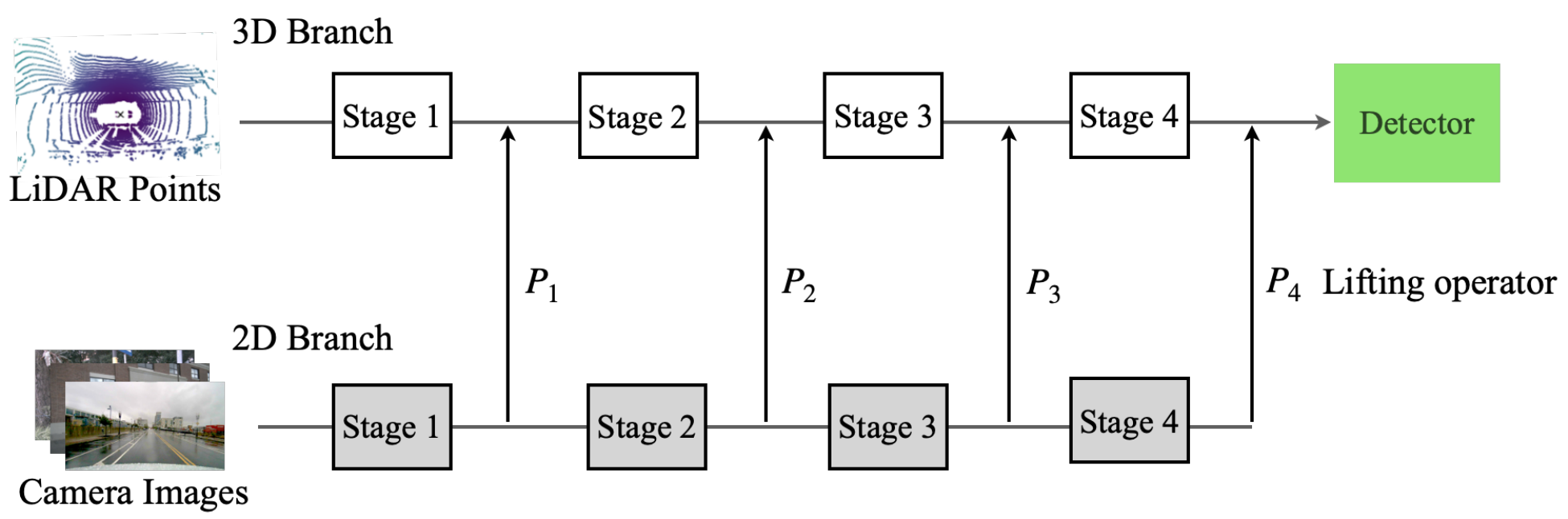}
    \caption{\blue{A generic 3D detection network with 2D feature fusion at different stages.}}
    \label{fig:network_overview}
\end{figure*}
\paragraph{3D Object Detection with LiDAR or Camera}
The goal of 3D object detection is to predict the 3D bounding box.
Methods based on LiDAR mainly involve encoding the raw LiDAR point clouds~\citep{pointnet, pointnet++}, or voxelizing them into sparse voxels~\citep{voxelnet}.
Building upon these 3D features, a variety of single-stage and two-stage 3D detection approaches~\citep{pvrcnn, point-rcnn, voxel-rcnn, deformable-pvrcnn, second, 3dssd, sassd, cia-ssd, pointpillars, centerpoint, second, part-a2, pillardet} have been proposed to predict the 3D bounding boxes of the target objects.

Another series of works focus on camera-based methods, wherein single-view or multi-view images are encoded using 2D backbones like ResNet~\citep{resnet}. Due to the absence of depth information, 2D-to-3D detection heads are developed to augment 2D features with implicitly or explicitly predicted depth, enabling the generation of 3D bounding boxes~\citep{liu2022petr, liu2022petrv2, detr3d, huang2021bevdet, reading2021categorical, xie2022m, huang2022bevdet4d, li2022bevformer, wang2021fcos3d}.

\paragraph{LiDAR-camera Fusion}
Due to the complementary properties of LiDARs and cameras, recent methods propose the joint optimization of both modalities, achieving superior accuracy compared to methods that rely solely on LiDAR or cameras. Illustrated in Figure~\ref{fig:overview} (a) and (b), these methods can be broadly categorized into two groups based on their fusion stages:
(a) Shallow fusion enhances the point clouds or shallow LiDAR features with image features, enriching the LiDAR inputs with image semantic priors~\citep{pointpainting, yin2021multimodal, xu2021fusionpainting, chen2022autoalign, chen2022autoalignv2, wu2022sparse, Li_2022_CVPR, focalsconv-chen, chen2022scaling, deepfusion1}.
(b) Deep fusion involves lifting image features into 3D space and integrating them in the middle or deep stages of the backbone network~\citep{liang2018deep, epnet}.

Shallow fusion methods, such as Focals Conv~\citep{focalsconv-chen} and LargeKernel3D~\citep{chen2022scaling}, have achieved state-of-the-art accuracy. In contrast, deep fusion models suffer from increasingly severe misalignment between camera and LiDAR features. Recently, Transfusion~\citep{bai2022transfusion} and DeepFusion~\citep{deepfusion} propose alignment of LiDAR and camera features using transformers and leverage cross-attention to dynamically capture correlations between image and LiDAR features during fusion. However, the computational cost of attention scales linearly with the product of the number of voxels and pixels, significantly increasing model inference latency. For instance, the attention module in DeepFusion~\citep{deepfusion} leads to over a 30\% increase in latency. To fully exploit the advantages of both shallow and deep-level fusion methods, our regularization-based approach provides a novel model training perspective, demonstrating that existing simple fusion methods can also handle the issue of semantic misalignment with appropriate regularization.

\paragraph{Path and Cycle consistency}
To promote feature alignment in our study, we incorporate a regularization term during training based on path consistency. Path and cycle consistency are common methods used to validate whether samples in a graph can match after traversing paths or cycles. In the deep learning community, these methods find application in image matching~\citep{zhou2015flowweb, zhou2016learning, zhou2015multi, dwibedi2019temporal}, domain transfer and adaptation~\citep{zhu2017unpaired, hoffman2018cycada, yi2017dualgan}, and enhancing feature quality in various general tasks, including segmentation and detection~\citep{zhang2019path, wang2013image, wang2014unsupervised}. Our work introduces path consistency into the extraction and fusion of LiDAR and camera features, marking the first instance of such integration to enable accurate 3D object detection.

\section{\blue{Background:} 2D and 3D feature fusion}
\begin{figure*}[!htbp]
    \centering
    \includegraphics[width=0.9\textwidth]{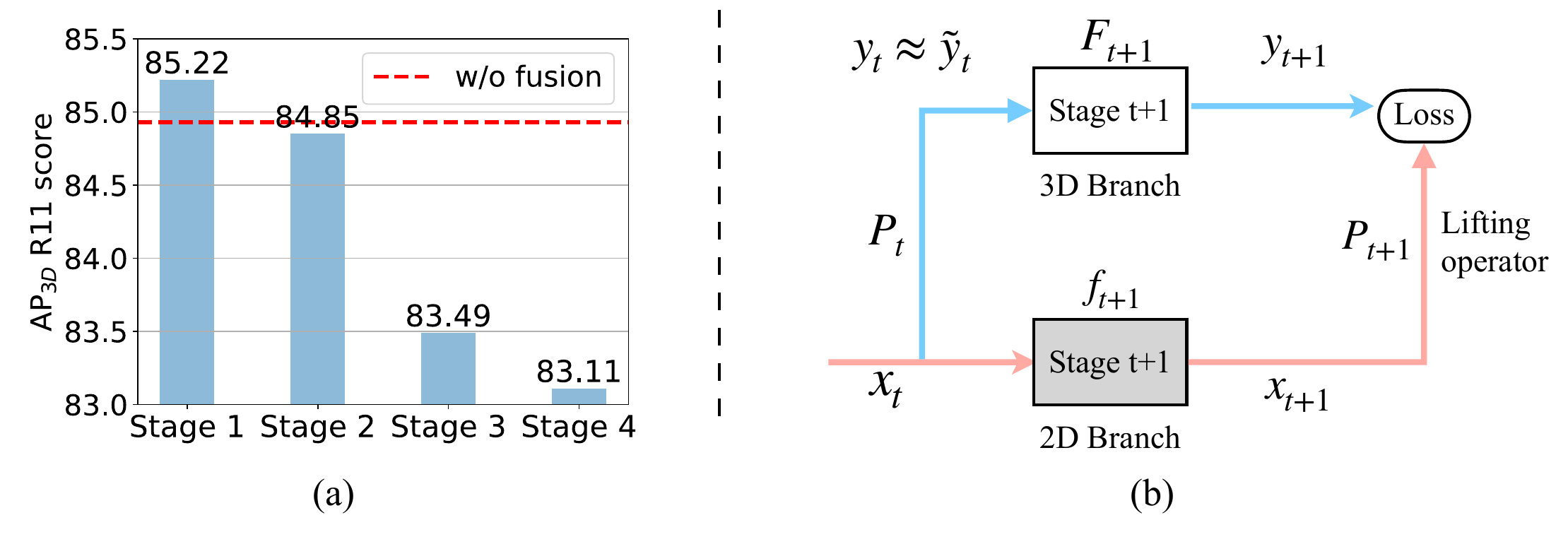}
\caption{(a) Illustration of performance degradation with naive deep feature fusion. Results are on the KITTI val split. The baseline setup without feature fusion achieves a 84.93\% of $\text{AP}_{\text{3D}}(R11)$. (b) Illustration of our path-consistent loss.}
\label{fig:pathloss}
\end{figure*}
The goal of Lidar-Camera feature fusion is to enhance 3D detection tasks with
additional dense 2D semantic information captured by cameras.
A generic 3D detection network with multi-stage 2D feature fusion is shown in Figure~\ref{fig:network_overview}. The network mainly consists of three components: a 3D feature extraction branch, a 2D feature extraction branch, and 2D-to-3D feature lifting. 

\paragraph{3D Feature Extraction} Given a point cloud containing a set of $N$ points, denoted as $\{p_i\}_{i=1}^N$ ($p_i\in \mathbb{R}^3$),
the objective is to detect objects of interest within the point cloud.
A typical preprocessing step involves voxelizing of the points, and transforming the representation of the point cloud into a voxel volume representation. Let $y_0 \in \mathbb{R}^{H\times W\times D}$ represent the resulting voxel volume, where $H$, $W$, and $D$ respectively correspond to the height, width, and depth within the 3D space.

The 3D detection network is built with a stack of a 3D feature blocks \citep[e.g.][]{pointnet++,voxelnet,voxeltransformer} and a detector head \citep[e.g.][]{pointpillars,pvrcnn,centerpoint,voxel-rcnn} with $y_0$ serving as the input.
The 3D backbone often has multiple stages, with down-sampling applied between each stage. 
Let $y_i \in \mathbb{R}^{C_i \times H_i\times W_i \times D_i}$ denote the feature maps generated by stage $i$, where $C_i$ represents the channel size and  $H_i, W_i, D_i$ represent the respective height, width and depth after the down-sampling.

\paragraph{\blue{2D Feature Extraction}} Assume a RGB image denoted as $x_0 \in \mathbb{R}^{3\times H'\times W'}$, which is captured along with the point cloud, where $H'\times W'$ represents the image resolution.
Similarly, using a modern feature backbone (such as ResNet50), 
one can extract 2D features for $x_0$ at different stages. Denote $x_i \in \R^{C_i'\times H_i' \times W_i'}$ the extracted 2D feature at stage $i$, where $C'_i$ denotes the channel size and $H_i' \times W_i'$ represents the size of the feature map.
The line of 2D and 3D feature fusion work aims at lifting the 2D features $\{x_i\}$ to the 3D feature space and fusing these lifted features with the 3D features $\{y_i\}$ to enhance 3D detection tasks.

\begin{figure*}[!tb]
    \centering
    \includegraphics[width=0.8\textwidth]{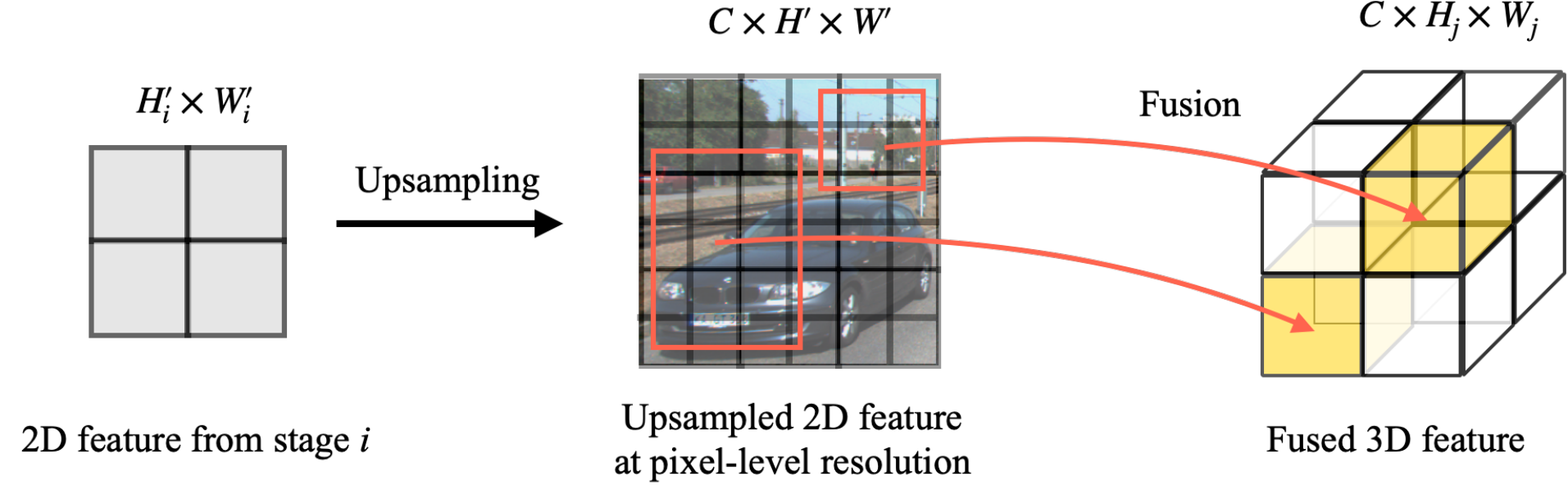}
    \caption{An illustration demonstrates the process of lifting features from 2D to 3D at a deeper stage. The upsampling is commonly implemented with a feature pyramid network~\citep{lin2017feature}. The image is sourced from~\citep{kitti}.}
    \label{fig:fusion_overview}
\end{figure*}

\paragraph{\blue{2D-to-3D Feature Lifting}}
An immediate challenge arises due to the distinct spatial resolutions of 2D and 3D feature maps, as well as their association with varying locations within the real-world 3D space. For illustrative purposes, let's consider the fusion of a 2D feature map $x_i$ from stage $i$ within the 2D branch, along with a 3D feature map $y_j$ from stage $j$ within the 3D branch.
Existing approaches~\citep[e.g.,][]{focalsconv-chen,deepfusion,chen2022scaling} often work by first upsampling $x_i$ to the original pixel-level resolution $H' \times W'$ using a feature pyramid network~\citep{lin2017feature}.
Then for each voxel within the 3D volume $y_i$, one can use the camera parameters to retrieve the corresponding 2D pixel-level feature for that voxel~\citep[e.g.,][]{pointpainting,deepfusion,bai2022transfusion,focalsconv-chen}. A demonstration of this process can be found in Figure~\ref{fig:fusion_overview}.
It's important to note that each voxel might correspond to a region of varying sizes. In this work, we follow the approach outlined in~\citep{focalsconv-chen} and take the feature vector at the center as the representation of the image patch for the sake of simplicity.
Finally, the fused 3D features could be a summation or concatenation of the lifted 2D features and the original 3D features derived from the 3D backbone.



\section{\blue{Motivating Example: Challenges in Deep Feature Fusion}}


Feature fusion at the input level is intuitive and straightforward, as each LiDAR point can accurately associate with its corresponding 2D pixels using the camera parameters. However, the precise mapping between 2D and 3D is compromised due to feature aggregation in both the 2D and 3D backbones. This becomes especially pronounced in deep layers, where each voxel corresponds to a broader spatial region. Each region encompasses pixels that are significantly distant from each other in the 3D space, as exemplified by the car and background trees in Figure~\ref{fig:fusion_overview}.

Ideally, one would seek to have each voxel and its corresponding 2D feature patch encode the same aspect of the physical world. Nonetheless, this correlation cannot be guaranteed within the network, given that both 2D and 3D features undergo non-linear transformations and down-sampling.
The consequence of this misalignment becomes more apparent as stages progress deeper and query regions expand. 
Naively attributing each voxel with 2D features that lack semantic relevance might lead to confusion within the 3D backbone, consequently compromising the performance of 3D detection.

We verify our hypothesis in Figure~\ref{fig:pathloss} (a). 
We closely follow the setting in prior-art Focals Conv~\citep{focalsconv-chen}, and alternate feature fusion from stage 1 to stage 4 as shown in Figure~\ref{fig:network_overview}. 
In this case, we use VoxelNet~\citep{voxelnet} with
Focals Conv as the 3D backbone and ResNet-50~\citep{resnet} as the 2D backbone. Both the 2D and 3D backbones have four stages. 
We report $\text{AP}_{\text{3D}}$ (R11) score on KITTI~\citep{kitti}. 
As shown in Figure~\ref{fig:pathloss} (a), 
the model trained with shallow feature fusion at stage 1 outperforms the baseline with no fusion, demonstrating the benefits of camera features for 3D detection.
However, the performance drops consistently when fusing features at deeper layers due to potential semantic misalignment between 2D and 3D features.
We propose a novel training regularization to address this issue in the following section.




\section{\blue{Method:} Deep Fusion with Path Consistency Regularization}




We denote $x_i \sim y_i$ as an ideal semantic alignment between 2D features $x_i$ and 3D features $y_i$, in a way that there exists a lifting operator $P_i$, such that the lifted 2D features $\tilde{y}_i = P_i \circ x_i$ and $y_i$ are semantically aligned.
We denote $\tilde{y}_i \approx y_i$ if $\tilde{y}_i$ and $y_i$ are semantically aligned, in a sense that $\tilde{y}_i$ and $y_i$ has a high similarity in the feature space.

For feature fusion between camera and LiDAR, the relationship 
$x_0 \sim y_0$ is valid at the input layer, where the mapping from 2D to 3D is exact. And $P_0$ can be constructed accurately with camera parameters.
Let's assume $x_t \sim y_t$ holds for features at stage $t$. This implies there exists a $P_t$ such that $\tilde{y}_t=P_t \circ x_t \approx y_t$.
It remains to show $x_{t+1}\sim y_{t+1}$ so that we can conclude features at all stages are semantically aligned from the perspective of mathematical induction.

In the following, we first show how $x_t$ and $y_t$ is transformed in the 2D branch and 3D branch, respectively. 
Subsequently, we present a path-consistent loss that encourages the network to learn
semantically aligned 2D and 3D features, that is $x_{t+1} \sim y_{t+1}$ (or $P_{t+1} \circ x_{t+1} \approx y_{t+1}$).


\paragraph{2D path}
On the 2D branch, $x_t$ is first evolved with a stack of 2D convolutions layers (denoted as $f_{t+1}$) and then lifted by $P_{t+1}$.
More precisely, the resulting 3D features can be written as the following, 
\begin{equation}
    y_{2D} = P_{t+1} \circ x_{t+1} =  P_{t+1} \circ f_{t+1} \circ x_t.
    \label{eq:path2d}
\end{equation}
Here, $x_{t+1}$ corresponds to the output derived from $f_{t+1}$.
We denote $A\circ B$ as applying operator $A$ on the output from $B$.
See the red path depicted in Figure~\ref{fig:pathloss} (b) for an illustration, 

\paragraph{3D path}
The 3D counterpart, denoted as $y_t$, undergoes a straightforward transformation using a series of 3D convolution layers, represented as $F_{t+1}$),
\begin{equation}
    y_{3D} = F_{t+1} \circ y_t \approx F_{t+1} \circ \tilde{y}_t = F_{t+1} \circ P_t \circ x_t.
    \label{eq:path3d}
\end{equation}
Here we make the substitution $y_t\approx \tilde{y}_t = P_t \circ x_t$ based on our construction.
See the blue path in Figure~\ref{fig:pathloss} (b) for an illustration.

\paragraph{Path-consistent regularization}
Our path-consistent regularization is simply defined as minimizing the distance between the lifted features $y_{2D}$ and the 3D features $y_{3D}$ such that $x_{t+1}\sim y_{t+1}$ as follows, 
\begin{align}
  & \mathcal{L}_{\text{consistency}}^t = \emph{loss}(y_{2D}, ~y_{3D}) \nonumber \\
  &  = {\emph loss}( \underbrace{P_{t+1} \circ f_{t+1} \circ x_t}_\text{2D path}, ~~\underbrace{F_{t+1} \circ P_t \circ x_t}_\text{3D path} ),
\label{eq:pathloss}
\end{align}
where ${\emph loss(\cdot)}$ denotes the loss function. 

Note that our consistency loss only depends on the same input $x_t$.
The information from both paths is dense. 
Furthermore, a simple loss function like negative cosine similarity or $\ell_1$ distance works well because the features come from the same camera feature space.

An alternative implementation of the consistency regularization is to minimize 
the difference between features extracted from camera data and Lidar data, 
that is minimizing ${\emph loss}(P_{t+1} \circ f_{t+1} \circ x_t, ~F_{t+1} \circ y_t)$.
However, in practice, the 3D voxel feature $y_t$ is often much sparser in comparison to the 2D feature $x_t$. This loss function is only applicable in region where $y_t$ has values.
Meanwhile, $F_{t+1}\circ y_t$ represents geometry-based features
while $P_{t+1} \circ x_{t+1}$ represents color-based features.
Consequently, designing a suitable loss function to measure the distance between these two is a complex task.

\paragraph{Path consistency for multi-stage networks} 
In the case of mutli-stage feature fusion, where features are fused at both 
shallow and layers. 
Assume a case with a total of $n$ stages. We apply the path consistency as defined in Eqn.~\ref{eq:pathloss} to all these stages, that is, 
\begin{align}
        & \mathcal{L}_{\text{consistency}} = \sum_{t=1}^{n-1}\mathcal{L}_\text{consistency}^t \nonumber \\ & =   \sum_{t=1}^{n-1} {\emph loss}( \underbrace{P_{t+1} \circ f_{t+1} \circ x_t}_\text{2D path}, 
        ~~~\underbrace{F_{t+1} \circ P_t \circ x_t}_\text{3D path}).
    ~\label{eq:pathloss_multi}
\end{align}

We refer to Eqn~\ref{eq:pathloss_multi} as our path consistency loss.






\paragraph{Algorithm} 
We augment the standard 3D detection loss with our path consistency loss. 
Overall, our training objective can be formulated as follows,
\begin{equation}
    \mathcal{L} = \mathcal{L} _{3D} + \alpha \mathcal{L}_{\text{consistency}}.
    \label{eq:loss}
\end{equation}
Here $\mathcal{L}_{3D}$ is the primary loss for the 3D task that affects all the parameters. $\mathcal{L}_{\text{consistency}}$ represents our regularization loss for ensuring path consistency, with $\alpha$ functioning as a hyper-parameter that determines the extent of this regularization's impact.
As our goal is to regularize the 2D branch to improve the 3D detection task, we only back-propagate our path consistency loss through the lifting operators and the 2D backbone while stopping the corresponding gradients on the 3D backbone.

\section{Experiments}

\begin{table*}[!hpbt]
\begin{center}
 \setlength{\tabcolsep}{6.5mm}
\begin{tabular}{l|c|ccc}
\Xhline{3\arrayrulewidth} 
{Method}  &   {Modality}   & Easy    & Moderate & Hard          \\ \hline 
 
3DSSD~\citep{3dssd}  & L & 89.71  & 79.45     & 78.67         \\
STD~\citep{std}    & L & 89.70   & 79.80  & 79.30  \\
SA-SSD~\citep{sassd} &  L & 90.15  & 79.91 & 78.78  \\
PV-RCNN~\citep{pvrcnn}   & L & 89.35    & 83.69    & 78.70   \\
VoTr-TSD~\citep{voxeltransformer}   &   L& 89.04   & 84.04     & 78.68         \\
Pyramid-PV~\citep{pyramid-rcnn}    & L    & 89.37   & 84.38  & 78.84  \\
Voxel R-CNN~\citep{voxel-rcnn}   &   L   & 89.41   & 84.54   & 78.93   \\
Focals Conv~\citep{focalsconv-chen}    &   L & 89.52  & 84.93  & 79.18  \\
LargeKernel3D~\citep{chen2022scaling}    &   L & 89.52 & 85.07 & 79.32  \\
\hline 
\hline
F-PointNet~\citep{fpointnet}   & LC   & 83.76  & 70.92 & 63.65   \\ 
PointSIFT+SENet~\citep{pointsiftsenet}  &  LC  & 85.62     &   72.05    &    64.19      \\ 
3D-CVF~\citep{3dcvf}  & LC &   89.67   &  79.88    &    78.47   \\
Focals Conv-F~\citep{focalsconv-chen}   &   LC  & 89.82  & 85.22  & 85.19  \\
\cellcolor{Graylight}\textit{PathFusion (ours)}   & \cellcolor{Graylight}LC   & \cellcolor{Graylight} \textbf{90.23}  & \cellcolor{Graylight}\textbf{85.84}  & \cellcolor{Graylight} \textbf{85.31}  \\
\Xhline{3\arrayrulewidth} 
\end{tabular}
\caption{Comparison on KITTI. Results are AP$_{\textrm{3D}}$ (R11) for { Car} on the {val} split. 
L represents LiDAR input and LC represents both LiDAR and camera input. Focals Conv-F denotes the implementation with shallow fusion on stage 1 in Figure ~\ref{fig:network_overview}.}
\label{tab:kitti-val}
\end{center}
\end{table*}
We evaluate our method, PathFusion, on both the KITTI~\citep{kitti} dataset and the nuScenes~\citep{nuscenes} dataset.
We choose Focals Conv~\citep{focalsconv-chen} as our base model. 
We add our path-consistent loss on top of Focals Conv training to facilitate semantically aligned feature fusion between the 2D and 3D branches.
Notably, we found that naively fusing LiDAR and camera features at deeper layers, akin to Focals Conv, results in significant performance degradation (as seen in Figure~\ref{fig:pathloss} (a)). 
However, our path-consistent regularization learns semantically aligned 
transforms from 2D to 3D, yielding consistent improvements as deeper features are fused. 
Specifically, when compared to Focals Conv using the same detector Voxel-RCNN, 
we achieved improvements in  $\text{AP}_{\text{3D}}$ (R11) from 85.22\% to 85.84\% on KITTI, and
an improved mAP from 70.1\% to 71.3\% on nuScenes testing split.

\paragraph{PathFusion settings}

Focals Conv fuses stage 1 features from the 2D branch. We denote this implementation as Focals Conv-F. 
Building upon the Focals Conv-F setup, we introduce deep-level feature fusion using our path-consistent regularization.
As outlined in Eqn.~\ref{eq:pathloss_multi}, 
we only apply our path consistency loss from stages beyond stage 1 because the misalignment concern only becomes substantial in deeper layers.  

We choose cosine as our consistency loss and set $\alpha = 0.01$ in Eqn.~\ref{eq:loss} as default, , unless stated otherwise.
Furthermore, our consistency regularization impacts parameters in a stage-wise manner. To elaborate, the consistency loss computed at the conclusion of stage $i$ exclusively influences the back-propagation of parameters within stage $i$ and the corresponding lifting operator $P_i$.


\subsection{Results on KITTI}
\paragraph{Dataset}
We follow the setting on KITTI~\citep{kitti} by splitting the dataset into
3,717 examples for training and  3,769 examples for validation. Each  example in the KITTI dataset consists of a LiDAR input coupled with two camera inputs.
We only use the LiDAR input and a single RGB image from the left camera for a fair comparison with Focals Conv~\citep{focalsconv-chen}.

\paragraph{Settings}
Following the setup in Focals Conv~\citep{focalsconv-chen}, we use Voxel RCNN~\citep{voxel-rcnn} as as the overall detection framework, in which we use VoxelNet~\citep{voxelnet} with Focals Conv as the 3D backbone and ResNet-50~\citep{resnet} as the 2D backbone. 
The ResNet-50 is pretrained on COCO~\citep{mscoco} with Deeplabv3~\citep{deeplab}.
To enable deep-level fusion, we introduce feature fusion in stage 4 and keep the original shallow-level fusion in stage 1.


We closely follow the training setting utilized in Focals Conv-F and train the model for 80 epochs using a batch size of 16. We use Adam optimizer with a learning rate of 0.01.
We report the mean Average Precision (mAP) metric. The mAP calculates the 3D bounding box recall score on 11 positions. The metric is denoted as $\text{AP}_{\text{3D}}$(R11).

\paragraph{Results}
We summarize our results in Table~\ref{tab:kitti-val}.
We report the $\text{AP}_{\text{3D}}$(R11) scores evaluated across the easy, moderate and hard levels following the previous studies~\citep{focalsconv-chen,voxel-rcnn,pvrcnn}. 
As we can see from Table~\ref{tab:kitti-val}, our method outperforms Focals Conv-F by 0.41\%, 0.62\%, 0.12\% across the three levels and achieves $90.23\%$, $85.84\%$ and $85.31\%$ on $\text{AP}_{\text{3D}}$(R11). These results imply that our proposed PathFusion excels in learning  better feature fusion,  consistently leading to improved results.



\begin{table*}[!hbpt]
\begin{center}
\resizebox{\linewidth}{!}{
\begin{tabular}{l|cc|cccccccccc}
\Xhline{3\arrayrulewidth} 
{Method}  &  mAP   & NDS & Car  & Truck  & Bus    & Trailer              & C.V. & Ped  & Mot   & Byc & T.C.  & Bar\\ \hline 
Focals Conv-F & 63.8 & 69.4 & 86.5 & 58.5 & 72.4 & 41.2 & 23.9 & 86.0 & 69.0 & 55.2 & 76.8 & 69.1 \\  \hline
\textit{PathFusion (ours)}   & \textbf{65.3} & \textbf{70.1} & 86.8 & 61.4 & 72.1 & 42.3 & 26.6 & 87.0 & 75.2 & 61.0 & 77.5 & 66.2 \\ 
\Xhline{3\arrayrulewidth} 
\end{tabular}}

\caption{
Comparison with other methods on nuScenes {val} split \textit{without} testing-time augmentations. }
\label{tab:nuscenes-val}
\end{center}
\end{table*}

\begin{table*}[!htbp]

\begin{center}
 \setlength{\tabcolsep}{1.0mm}
    \renewcommand\arraystretch{1.2}
\resizebox{0.9\linewidth}{!}{
\begin{tabular}{l|c|cc|cccccccccc}
\Xhline{3\arrayrulewidth} 
 {Method} & {Modality} & mAP   & NDS & Car  & Truck   & Bus   & Trailer    & C.V. & Ped   & Mot  & Byc     & T.C.  & Bar   \\ \hline 
3DSSD~\citep{3dssd}                         & L & 42.6 & 56.4 &  81.2 & 47.2 & 61.4 & 30.5 & 12.6 & 70.2 & 36.0 & 8.6 & 31.1 & 47.9 \\ 
CBGS~\citep{cbgs}                        & L & 52.8 & 63.3 & 81.1 & 48.5 & 54.9 & 42.9 & 10.5 & 80.1 & 51.5 & 22.3 & 70.9 & 65.7 \\ 
HotSpotNet~\citep{hotspotnet}                   & L & 59.3 & 66.0 & 83.1 & 50.9 & 56.4 & 53.3 & 23.0 & 81.3 & 63.5 & 36.6 & 73.0 & 71.6 \\
CVCNET~\citep{cvcnet}                  &L  & 58.2 & 66.6  & 82.6 & 49.5 & 59.4 & 51.1 & 16.2 & 83.0 & 61.8 & 38.8 & 69.7 & 69.7 \\ 
CenterPoint~\citep{centerpoint}                         & L & 58.0 & 65.5 & 84.6 & 51.0 & 60.2 & 53.2 & 17.5 & 83.4 & 53.7 & 28.7 & 76.7 & 70.9 \\ 

CenterPoint$^{\dagger}$                         & L & 60.3 & 67.3 & 85.2 & 53.5 & 63.6 & 56.0 & 20.0 & 84.6 & 59.5 & 30.7 & 78.4 & 71.1 \\
Focals Conv~\citep{focalsconv-chen}                      & L & 63.8 & 70.0 & 86.7 & 56.3 & 67.7 & 59.5 & 23.8 & 87.5 & 64.5 & 36.3 & 81.4 & 74.1 \\ 
LargeKernel3D~\citep{chen2022scaling} &L & 65.3 & 70.5 & 85.9 &55.3 &66.2 &60.2& 26.8 &85.6 &72.5 &46.6& 80.0 &74.3 \\

\hline \hline
PointPainting~\citep{pointpainting}                         & LC & 46.4 & 58.1 & 77.9 & 35.8 & 36.2 & 37.3 & 15.8 & 73.3 & 41.5 & 24.1 & 62.4 & 60.2 \\ 
3DCVF~\citep{3dcvf}                        & LC  & 52.7 & 62.3 & 83.0 & 45.0 & 48.8 & 49.6 & 15.9 & 74.2 & 51.2 & 30.4 & 62.9 & 65.9 \\ 
FusionPainting~\citep{fusionpainting} &LC   & 66.3 & 70.4 & 86.3 & 58.5 & 66.8 & 59.4 & 27.7 & 87.5 & 71.2 & 51.7 & 84.2 & 70.2 \\ 
MVF~\citep{mvf}                         & LC  & 66.4 & 70.5 & 86.8 & 58.5 & 67.4 & 57.3 & 26.1 & 89.1 & 70.0 & 49.3 & 85.0 & 74.8 \\
PointAugmenting~\citep{pointaugmenting}                        & LC  & 66.8 & 71.0 & 87.5 & 57.3 & 65.2 & 60.7 & 28.0 & 87.9 & 74.3 & 50.9 & 83.6 & 72.6 \\
\hline

Focals Conv-F~\citep{focalsconv-chen}                     & LC  & 67.8 & 71.8 & 86.5 & 57.5 & 68.7 & 60.6 & 31.2 & 87.3 & 76.4 & 52.5 & 84.6 & 72.3 \\ 
\cellcolor{Graylight} \textit{PathFusion (ours)}                       & \cellcolor{Graylight}LC  & \cellcolor{Graylight} \textbf{69.0} & \cellcolor{Graylight} \textbf{72.6} & \cellcolor{Graylight} 87.5 & \cellcolor{Graylight} 59.8 & \cellcolor{Graylight} 69.3 & \cellcolor{Graylight} 62.0 & \cellcolor{Graylight} 34.7 & \cellcolor{Graylight} 87.6 & \cellcolor{Graylight} 77.3 & \cellcolor{Graylight} 53.7 & \cellcolor{Graylight} 85.3 & \cellcolor{Graylight} 72.9 \\ 
\hline
Focals Conv-F$^{\dagger}$                        &LC & 68.9 & 72.8 & 86.9 & 59.3 & 68.7 & 62.5 & 32.8 & 87.8 & 78.5 & 53.9 & 85.5 & 72.8 \\ 
\cellcolor{Graylight} \textit{PathFusion (ours)}$^{\dagger}$                      & \cellcolor{Graylight}LC   & \cellcolor{Graylight} \textbf{70.2} & \cellcolor{Graylight} \textbf{73.6} & \cellcolor{Graylight} 88.0 & \cellcolor{Graylight} 61.4 & \cellcolor{Graylight} 69.4 & \cellcolor{Graylight} 64.1 & \cellcolor{Graylight} 34.9 & \cellcolor{Graylight} 88.5 & \cellcolor{Graylight} 80.2 & \cellcolor{Graylight} 54.8 & \cellcolor{Graylight} 85.9 & \cellcolor{Graylight} 73.2 \\ 
\hline
Focals Conv-F$^{\ddagger}$                        &LC  & 70.1 & 73.6 & 87.5 & 60.0 & 69.9 & 64.0 & 32.6 & 89.0 & 81.1 & 59.2 & 85.5 & 71.8 \\
\cellcolor{Graylight} \textit{PathFusion (ours)} $^{\ddagger}$                       &  \cellcolor{Graylight}LC  & \cellcolor{Graylight} \textbf{71.3} & \cellcolor{Graylight} \textbf{74.3} & \cellcolor{Graylight} 88.7 & \cellcolor{Graylight} 62.6 & \cellcolor{Graylight} 70.1 & \cellcolor{Graylight} 64.9 & \cellcolor{Graylight} 35.9 & \cellcolor{Graylight} 89.8 & \cellcolor{Graylight} 82.5 & \cellcolor{Graylight} 59.6 & \cellcolor{Graylight} 86.0 & \cellcolor{Graylight} 72.4 \\ 
\Xhline{3\arrayrulewidth} 

\end{tabular}}
\caption{Comparison on the nuScenes test set. "L" represents LiDAR input, while "LC" represents a combination of LiDAR and camera input. The symbol $^{\dagger}$ indicates double flip, and $^{\ddagger}$ signifies double flip and rotation testing-time augmentation.
}
\label{tab:nuscenes-test}
\end{center}
\end{table*}

\subsection{Results on nuScenes}
\paragraph{Dataset}
nuScenes~\citep{nuscenes} \footnote{https://www.nuscenes.org/} is a large-scale 3D object detection dataset for self-driving.
It consists of 700 scenes for training, 150 scenes for validation and additional 150 scenes for testing.  
Each scene has a sequence of frames, with each frame containing data captured by one LiDAR sensor and six cameras. 
In total, nuScenes contains 1.4M camera images and 390K LiDAR sweeps.

\paragraph{Settings}
Following Focals Conv-F~\citep{focalsconv-chen}, we use VoxelNet as our 3D backbone and use ResNet-50 pretrained using DeeplabV3~\citep{deeplab} on COCO~\citep{mscoco} as the 2D backbone. We train the model for 20 epochs with batch size of 32 using Adam Optimizer with a learning rate of 0.001. We apply a cosine scheduler to decrease the learning rate to $1e^{-4}$. For the data augmentation, same as Focals Conv-F, we apply rotation, flip, translate, rescale and ground truth sampling augmentation during training. 
We extend the Focals Conv-F to fuse 2D and 3D features at both stage 1 and stage 2, see Figure~\ref{fig:network_overview}.

\paragraph{Results} 
We evaluate on both the nuScenes validation and test set, and report our results in Table~\ref{tab:nuscenes-val} and Table~\ref{tab:nuscenes-test}, respectively.
On the validation split, 
we achieve a $1.5\%$ improvement on mAP and a $0.7\%$ improvement on NuScenes Detection Score (NDS) compared to Focals Conv-F. 

We further evaluate our model on the nuScenes test server.
In line with~\citet{liu2022bevfusion,liang2022bevfusion,focalsconv-chen}, we report results with and without test-time data augmentation. 
For the testing-time augmentations, we follow ~\cite{focalsconv-chen} and use double flip and rotation with yaw angles spanning $[-6.25^{\circ}, 0^{\circ}, 6.25^{\circ}]$. 
We observe a consistent improvements over Focals Conv across all the settings, resulting from the incorporation of our path-consistent loss during training.
Specifically, our best result outperforms Focals Conv-F by 1.2\%, 1.3\%, 1.2\% mAP, and by 0.8\%, 0.8\%, 0.7\% NDS on the test set for the settings both without and with testing-time data augmentation.


\begin{table*}[!hbpt]
    \centering
         \setlength{\tabcolsep}{0.9mm}
    \renewcommand\arraystretch{1.1}
    \resizebox{\linewidth}{!}{
    \begin{tabular}{l|c|cccc|ccc|c}
   \Xhline{3\arrayrulewidth} 
        Method  & w/o Fusion & Stage 1 & Stage 2 & Stage 3  & Stage 4 & Stage 1\&2  & Stage 1\&3 & Stage 1\&4  & All Stages \\ \hline
        w/o path consistency & 84.93 & 85.22 & 84.85  & 83.49 & 83.11 & 85.01 &  84.85&   84.02 & 83.97 \\  
        w/ path consistency & - &  - & 
 \bf{85.69}  &  \bf{85.76}  &  \bf{85.78}&  \bf{85.79} & \bf{85.82} & \bf{85.84} & \bf{85.88}  \\ \hline \hline
        \emph{Relative improvements} & - & - & 0.84 & 2.27 & 2.67 &0.78 & 0.97 & 1.82 & 1.91 \\
  \Xhline{3\arrayrulewidth} 
    \end{tabular}}
    \caption{Ablation study on various fusion configurations. Results are $\text{AP}_{\text {3D}}$(R11) on the KITTI validation set. All Stages signifies the fusion from Stage 1 to Stage 4.}
    \label{tab:ablation_stage}
\end{table*}

\subsection{Ablation Studies}


\paragraph{Choices of fusion configurations}
To study the importance of feature fusion at deep layers, 
we vary the feature fusion at different stages. 
We first evaluate on the KITTI dataset. As depicted in Table~\ref{tab:ablation_stage},  without our path-consistent regularization, 
the best detection result is achieved when only the feature from stage 1 is fused. 
In contrast, our method leads to consistent improvements. And our best results are achieved by fusing all the stages.

We further evaluate on the nuScenes dataset. 
Here, to expedite experiment turnover, 
we adopt the practice of Focals Conv~\citep{focalsconv-chen} and only train on a $\frac{1}{4}$ subset of the complete training set.
As we can see from Table~\ref{tab:improvements-multimodal-1/4},
without our path-consistent regularization, both the mAP and NDS score decrease significantly when deep features are fused.
This trend aligns with our observations on the KITTI dataset. 
However, our method learns better feature fusion, resulting in a 0.7\% improvement in mAP and a 0.6\%
improvement in NDS.



\begin{table}[!bhpt]
\begin{center}
\begin{tabular}{l|ll}
\Xhline{3\arrayrulewidth} 
& mAP   & NDS                  \\ \hline
Shallow-level fusion   &  61.7                      & 67.2     \\
Deep-level fusion  & 60.7                       & 65.6     \\
 Shallow \& Deep level fusion   & 61.5                       & 67.0     \\
\cellcolor{Graylight} ~~~~~ + path consistency   & \cellcolor{Graylight}\textbf{62.4}    & \cellcolor{Graylight}\textbf{67.8}     \\
\Xhline{3\arrayrulewidth} 
\end{tabular}
\caption{Improvement over the baseline trained on a $\frac{1}{4}$ subset of nuScenes, evaluated on the validation set. }
\label{tab:improvements-multimodal-1/4}
\end{center}
\end{table}

\begin{table}[!bhpt]
    \centering
    \begin{tabular}{c|cccc}
    \Xhline{3\arrayrulewidth} 
        Method &  Mean & Min & Max\\ \hline
        Focals Conv-F &  $85.06\pm  0.34$ & 84.59 & 85.43\\
       PathFusion (ours) & $85.84\pm0.13$ & 85.70 & 86.07 \\
       \Xhline{3\arrayrulewidth} 
    \end{tabular}
    \caption{Comparison on KITTI through 10 random trials. Results are $\text{AP}_{\text{3D}}$ (R11) at the moderate level. \emph{Min} and \emph{Max} denotes 
    the lowest and highest $\text{AP}_{\text{3D}}$ (R11) score achieved out of the 10 runs.}
    \label{tab:kitti_robustness}
\end{table}

\paragraph{Robustness of our method}
We found our method leads to significantly improved training stability. Specifically, we use the official Focals Conv-F repo~\footnote{https://github.com/dvlab-research/FocalsConv} and repeat the experiments 10 times with different random seeds. We report the $\text{AP}_{\text{3D}}$(R11) at the moderate level in Table~\ref{tab:kitti_robustness}.
Compared with Focals Conv-F, 
our method yields a small standard deviation (i.e., 0.13). Notably, 
even our least favorable run outperforms the  best-performing run of Focals Conv-F. 
We report the average performance of our method in Table~\ref{tab:kitti-val}.

\paragraph{Impact of gradient stopping on 3D branch}
The practice of gradient stopping is a important trick for effectively regulating the corresponding parameters of the path consistency loss.
This importance stems from two key reasons:
1) Avoiding trivial solutions: Excluding gradient stopping would allow the path-consistency alone to generate trivial solutions, where both the 2D and 3D branches yield zero activations.
2/ Task focus: As our primary concern is 3D detection, it's logical to solely regulate the 2D branch to facilitate the 3D task, all while not interfering with the 3D branch.
To validate this, we conducted an experiment where we removed gradient stopping from the 3D backbone and assessed performance on the $\frac{1}{4}$ subset of the nuScenes dataset. Remarkably, we observed a significant drop in both mAP and NDS, from 62.4\% and 67.8\% to 56.7\% and 63.0\%, respectively. This substantiates the efficacy of employing gradient stopping on the 3D branch.


\paragraph{Consistency loss design}
We conduct experiments using both the cosine and ${\ell_1}$ loss as our 
consistency loss measures. 
And in the meantime, we study the impact of our consistency coefficient $\alpha$, presenting our findings in  Table~\ref{tab:loss_design} using the 1/4 nuScenes dataset.
As depicted in Table~\ref{tab:loss_design},
our method outperforms the baseline (mAP 61.7\% and NDS 67.2\%) across all settings. 
The best performance is achieved when employing the cosine loss with a coefficient of $\alpha=0.01$.

\begin{table}[!bhpt]
\begin{center}
\begin{tabular}{ll|cc}
\Xhline{3\arrayrulewidth} 
Loss Type & Loss weight ($\alpha$) & mAP   & NDS    \\ \hline
\multirow{3}{*}{Cosine} & 0.1    & 62.1  & 67.5     \\
& 0.01    & \textbf{62.4}  & \textbf{67.8}     \\
& 0.001    & 62.3  & 67.7     \\
\hline
\multirow{3}{*}{$\ell_1$} & 0.1 &  61.8 & 67.3     \\
& 0.01 &  62.0 & 67.5     \\
& 0.001 &  62.0 & 67.4     \\
\Xhline{3\arrayrulewidth} 
\end{tabular}
\caption{Ablation study on the impact of various path-consistent loss designs and the effect of the loss coefficient $\alpha$. The results are derived from training on the $\frac{1}{4}$ subset of the nuScenes dataset and evaluating on the validation set.}
\label{tab:loss_design}
\end{center}
\end{table}


\section{Conclusion}

In this work, we proposed a path consistency loss to
improve the feature fusion between LiDAR features and camera features at deeper layers.
Our method works by encouraging the 2D branch to follow the transformations learned 
in the 3D branch. This, in turn, generates complementary information that aligns semantically with the 3D features.
We applied our method to improve prior-art Focals Conv. As a result,  
our method leads to a large improvement on both the KITTI and nuScenes datasets.
Notably, our PathFusion achieves a mAP of 71.3\%  on the nuScenes test set, surpassing the Focals Conv-F by 1.2\%.



{
    \small
    \bibliographystyle{ieeenat_fullname}
    \bibliography{paper}
}

\end{document}